\documentclass[11pt]{article}

\usepackage[final]{acl}

\usepackage{times}
\usepackage{latexsym}

\usepackage[T1]{fontenc}
\usepackage[utf8]{inputenc}

\usepackage{microtype}

\usepackage{inconsolata}

\usepackage{graphicx}

\usepackage{subcaption}
\usepackage{enumitem}
\usepackage{amsmath}
\usepackage{hyperref}
\usepackage{xurl}
\usepackage{xcolor}
\usepackage[textsize=tiny]{todonotes}
\usepackage{enumitem}
\usepackage{booktabs}
\usepackage{multirow}

\definecolor{boxteal}{HTML}{B7E2DA}
\definecolor{inlineteal}{HTML}{91C4C3}

\usepackage[most]{tcolorbox}
\definecolor{bggray}{rgb}{0.95, 0.95, 0.95}
\usepackage[%
    framemethod=tikz,
    skipbelow=\topskip,
    skipabove=\topskip
]{mdframed}
\mdfsetup{%
    leftmargin=0pt,
    rightmargin=0pt,
    backgroundcolor=bggray,
    middlelinecolor=black,
    roundcorner=3
}
\newtcolorbox[list inside=prompt,auto counter,number within=section]{prompt}[1][]{
    colbacktitle=black!60,
    fonttitle=\small,
    coltitle=white,
    fontupper=\footnotesize,
    boxsep=4pt,
    left=0pt,
    top=0pt,
    bottom=0pt,
    boxrule=1pt,
    #1,
}

\title{On the Role of Citations in Preference Data}

\author{
 \textbf{Yu Hou\textsuperscript{1}},
 \textbf{Hal {Daum\'e III}\textsuperscript{1}},
 \textbf{Rachel Rudinger\textsuperscript{1}},
 \textbf{William Walden\textsuperscript{2}}
\\
\\
 \textsuperscript{1}University of Maryland,
 \textsuperscript{2}Johns Hopkins University
\\
 \small{
   \textbf{Correspondence:} \href{mailto:email@domain}{houyu@umd.edu,wwalden1@jh.edu}
 }
}

\newif\ifnotes
\notestrue % display notes
\definecolor{blue-d}{HTML}{0118D8}
\definecolor{teal-d}{HTML}{48A6A7}
\definecolor{purple-d}{HTML}{9112BC}
\definecolor{green-d}{HTML}{3CB371}

\begin{document}
\maketitle
\begin{abstract}
Many NLP tasks require systems to provide \emph{attribution} in their outputs---i.e.\ citations to grounding sources. Attribution serves as a bulwark against model hallucination and as a means for users to verify the credibility of model outputs. Yet, it is unclear how humans and LLMs evaluate citations when comparing outputs, a process central to reward modeling and modern LLM post-training.
This paper studies the role of citations in the preferences of human judges and LLMs within the context of scientific question answering, leveraging mixed effects models to investigate the influence of citations on pairwise judgments.
Among our key findings are (1) that humans prefer more \emph{diverse} citations but fewer citations \emph{overall}, and (2) that LLMs show some stronger citation-related preferences compared to humans, despite lacking access to the sources, but these preferences depend on the data and specific models. Finally, we discuss the implications of our findings for preference data collection.\footnote{Code and data can be found at: \url{https://github.com/houyu0930/citation-preference}.}

\end{abstract}

\section{Introduction}

Pairwise preference data plays an essential role in LLM development, both in post-training~\cite{ouyang2022traininglanguagemodelsfollow, li2025prefpalette} and in evaluation~\cite{chatbotarena}, enabling models to learn difficult-to-specify behaviors and to be efficiently ranked and evaluated.
But we pay for these benefits with interpretability: the reason for a given preference is generally not known---perhaps even to the assessors themselves. Such interpretability is vital to understanding dimensions of human and model bias, which in turn is a necessity for the development of fair and robust reward models~\cite{dubois2024lengthcontrolled, wu-etal-2025-rewordbench, bharadwaj2025flatteryflufffogdiagnosing, zhao2025tokenfoolllmasajudge, gehrmann2025rewardmodelsmetricstrench}.

\begin{figure}[t]
\centering
\includegraphics[width=\linewidth]{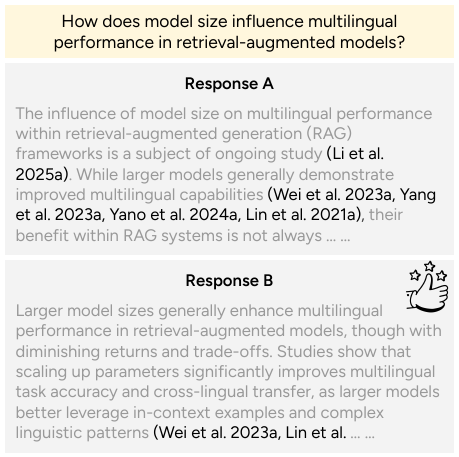}
\caption{A pair of citation-attributed model responses to a question from SciArena~\cite{zhao2025sciarenaopenevaluationplatform}. We use mixed-effects models to understand the impact of citations on human and LLM preferences on such pairs.%\vspace{-5mm}
}
\label{fig:example}
\end{figure}

% https://docs.google.com/drawings/d/1DBPM0eNBrbIZR23CJjsiQxQrxTQ3zZOwHbn7L8NImQc/edit?usp=sharing

This need has motivated a budding line of research on inferring interpretable explanations for preference judgments. Much of this work adopts a ``propose-and-validate'' paradigm, in which candidate explanations are \emph{proposed}---e.g.\ via LLM analysis~\cite{findeis2025InverseConstitutionalAI, zhong-etal-2024-explaining} or via sparse autoencoder features~\cite{movva2025whatshumanfeedbacklearning}---and then \emph{validated} by attempting to predict (human or model) judgments from these explanations. This work, though valuable, neglects a category of data core to most retrieval-augmented generation (RAG) and deep research applications---namely, long-form responses \emph{with citations}.

Specifically, as more people turn to AI as scientific research assistants---for finding papers or answering scholarly queries---citations play an increasingly important role in information-seeking tasks than they once did with traditional web search. In this work, we ask: \textbf{how do citation-related features influence human preferences in scientific question answering?}
Further, because LLM judges are often used for preference labeling as a proxy for human expert annotations, we compare: \textbf{how do LLMs weigh citations in their pairwise assessment of responses?}

% It is taken as given that proper citation is integral to any credible response to complex information-seeking queries. But is it? Or more precisely: What does preference data tell us about how human and LLM judges weigh citations in their assessment of response quality, controlling for other variables?

We draw on \emph{mixed-effects models}~\citep[MEMs;][]{gelman_data_2014} to interrogate these questions on two datasets, which have pairwise judgments on citation-backed model responses to diverse scientific questions (\autoref{fig:example}): SciArena~\citep{zhao2025sciarenaopenevaluationplatform} and ResearchQA~\citep{yifei2025researchqaevaluatingscholarlyquestion}.
We use human preferences from these datasets and set up LLM judge experiments to compare the observed effects on each dataset.
Specifically, we develop a structured approach for auditing preference data: we decompose judgments into interpretable dimensions and estimate their associations with human and LLM preferences separately, allowing a side-by-side comparison.
This approach is not specific to the datasets studied here, and can be applied to other preference collections.

% structured understanding of preference data
% I think the paper slightly undersells its most durable contribution, which is the structured approach for auditing preference data by decomposing judgments into interpretable dimensions and estimating their association with human and LLM preferences. The paper designed and implemented this strategy well, but it may benefit readers if the methodological contribution is foregrounded a bit more.

% a large collection of human and LLM preferences from SciArena~\citep{zhao2025sciarenaopenevaluationplatform}, a dataset of pairwise judgments on citation-backed model responses to diverse scientific questions (\autoref{fig:example}), .

% as well as on model attributes~\cite{dunlap_vibecheck}.
% Whereas these works seek to \emph{discover} interpretable and preference-relevant features, we probe a specific and 

% While those approaches are promising, their main focus is freeform preference discovery, explaining responses that do not have citations, where citations are returned from Retrieval Augmented Generation systems.
% Given the role of citations from disparate perspectives, the proposing step can lead to explanations that are highly correlated with each other, for example, ``response contains more citations'' and ``response contains more unique citations'', making the validating step less reliable.

Overall, we find that: \textbf{(1)} human judges tend to prefer responses with \emph{more diverse} but \emph{fewer overall} citations.
% On ResearchQA, we also find that humans do not prefer responses whose citations mismatch with the reference list.
\textbf{(2)} Experiments on four open-source LLMs (Llama, Gemma, Qwen, and Skywork) show that the influence of citation-related and non-citation features on LLM preferences differs from their influence on human preferences, highlighting potential LLM biases. 
\textbf{(3)} Among the four LLM judges, we find that they exhibit dissimilar correlational patterns and citation preferences.
We discuss the implications of these findings, hoping to motivate more careful attention to the treatment of citations in preference data collection.

\section{General Setup}
\label{sec:setup}

Although both SciArena~\citep{zhao2025sciarenaopenevaluationplatform} and ResearchQA~\citep{yifei2025researchqaevaluatingscholarlyquestion} focus on scientific QA scenarios, their processes for collecting preferences differ.
We discuss the high-level setup here and provide more details in Section~\ref{sec:results-sciarena} and Section~\ref{sec:results-researchqa}.

\paragraph{Dataset Selection Criteria} Our analysis requires datasets that satisfy three conditions. First, responses must contain citations.\footnote{We found that many preference datasets that \emph{ought} to satisfy this first condition do not.} Second, each query must have multiple responses so that pairwise comparisons are possible. Third, because we wish to compare LLM judge behavior with that of humans, each response pair must have a clear human preference vote, allowing us to probe LLM judge preferences on the same data. Following an expansive search of existing datasets, these requirements constrained us to only two options: SciArena, with $12,249$ comparisons, and ResearchQA, with $169$ available comparisons.
In both datasets, human annotators had access to the cited sources.

% \begin{figure}[t]
%     \centering
%     \begin{prompt}[title={LLM Judge Prompt}]
%     \texttt{You are an expert in scientific literature synthesis. Your task is to evaluate the quality of two AI-generated citation-attributed responses to a user's question. Assess both responses for relevance, accuracy, clarity, and appropriate use of citations. Then, select the response, Output (a) or Output (b), that best addresses the user's question.} \\ \\
%     \texttt{User Question:} \\
%     \texttt{\{question\}} \\ \\
%     \texttt{Output (a):} \\
%     \texttt{\{response\_a\}} \\ \\
%     \texttt{Output (b):} \\
%     \texttt{\{response\_b\}} \\ \\
%     \texttt{Which is best, Output(a) or Output (b)?} \\
%     \texttt{Answer with only: "A" or "B"}
%     \end{prompt}
%     \caption{LLM judge prompt used for the inference experiments. It is the same prompt used in SciArena.}
%     \label{fig:llm-judge-prompt}
% \end{figure}

\paragraph{LLM Inference Experiment} We elicit preferences on these examples from four LLMs: Llama3.3-70B~\citep{grattafiori2024llama3herdmodels}, Gemma3-27B~\citep{gemmateam2025gemma3technicalreport}, Qwen3-30B~\citep{yang2025qwen3technicalreport}, and Skywork-Critic-70B~\citep{skyworkcritic2024}. Skywork is specifically trained as a generative reward model and was selected due to its top rank on RAG-RewardBench~\citep{jin-etal-2025-rag}, while the other three were selected due to their general popularity among open-weight models, though they were not trained explicitly as reward models. % that excel across different tasks.
To control for biases arising from presentation order, we run experiments for both possible orderings of responses in the prompt, yielding $12,249 \times 2 = 24,498$ and $169 \times 2 = 338$ judgments per LLM for SciArena and ResearchQA, respectively.
We use the same judge prompt for both datasets 
(see below).
% (see Figure~\ref{fig:llm-judge-prompt}).
\footnote{For further experiment setup details, see \autoref{app:llm-exps}.}

\begin{prompt}[title={LLM Judge Prompt}]
\texttt{You are an expert in scientific literature synthesis. Your task is to evaluate the quality of two AI-generated citation-attributed responses to a user's question. Assess both responses for relevance, accuracy, clarity, and appropriate use of citations. Then, select the response, Output (a) or Output (b), that best address the user's question.} \\ \\
\texttt{User Question:} \\
\texttt{\{question\}} \\ \\
\texttt{Output (a):} \\
\texttt{\{response\_a\}} \\ \\
\texttt{Output (b):} \\
\texttt{\{response\_b\}} \\ \\
\texttt{Which is best, Output(a) or Output (b)?} \\
\texttt{Answer with only: "A" or "B"}
\end{prompt}

% \footnote{\autoref{app:llm-exps} contains the judge prompt and further experimental details. Controlling for presentation order is motivated by our early observations of skewed LLM preference distributions (e.g.\ Llama tends to pick the first option). This control is applied only to LLM judgments because the human preference data lacks information on presentation order.}

\paragraph{Mixed-Effects Models} (MEMs) are hierarchical regression models that incorporate slope and/or intercept parameters that vary by group (\emph{random effects}) as well as parameters that are shared across groups (\emph{fixed effects}). MEMs are widely used for interpretable data analysis \cite[][]{GelmanHill_2007, mcelreath_statistical_2018, federico-etal-2014-assessing, gantt-etal-2020-natural, white-etal-2022-mixed, sun-etal-2023-dialect, sandoval-etal-2023-rose, benito-santos-etal-2025-robust}. Following the Bradley-Terry model~\cite{bt-model}, we treat binary preferences as the dependent variable ($\boldsymbol{y}$) in a mixed-effects logistic regression:
$$
\text{logit}(P(\boldsymbol{y} = 1 \mid \mathbf{u})) = \mathbf{X}\boldsymbol{\beta} + \mathbf{Z}\mathbf{u}
$$
where $\mathbf{X}$ is a matrix of predictors; $\boldsymbol{\beta}$ is a vector of fixed-effects regression coefficients; $\mathbf{Z}$ is the design matrix for random effects; $\mathbf{u}$ is a vector of random effects, where $u_j \sim \mathcal{N}(\mathbf{0}, \mathbf{G})$, with $\mathbf{G}$ as the covariance matrix; and $\boldsymbol{\beta}$ and $\mathbf{G}$ are the parameters to be inferred.
% \footnote{Our mixed-effects model is a variation of the Bradley-Terry model. However, because our goal is to explain preferences rather than estimate LLM strength, we incorporate the models and related interactions as random effects.}

\paragraph{Fixed-Effects Predictors} We consider three citation-related variables for each dataset: citation diversity, citation count, citation age (SciArena only), and citation mismatch (ResearchQA only). These variables are a subset of those studied in citation bias research~\citep{wahle-etal-2025-citation, algaba-etal-2025-large, fang-etal-2025-llm-recency}, targeted at comparison of closed sets of citations rather than (fully open) citation selection (as responses in a pair are conditioned on the same query and retrieved documents).

\begin{description}[itemsep=0pt, parsep=1pt]
\item[Citation diversity ($\text{C}_D$)] captures the diversity of cited sources and is measured by the Shannon diversity index (i.e.\ entropy):
$-\sum_{i=1}^{n} p_i \log_2(p_i)$, where $n$ is the number of unique sources in the response and $p_i$ represents the proportion of citations attributed to source $i$. Low values indicate low diversity (i.e.\ when a response is dominated by only a small number of sources, repeatedly cited).
\item[Citation count ($\text{C}_C$)] measures the number of citations in a response. $\text{C}_C$ and $\text{C}_D$ are used to test hypotheses about citations' \emph{distribution}.
\item[Citation age ($\text{C}_A$)] \textbf{[SciArena only]} captures the average publication date of cited sources as presented in~\autoref{fig:example}, calculated as $\text{avg}(\ln(2026 - y))$, where $y$ ranges over citations' publication years.\footnote{We apply a log transformation as the date distribution is heavily skewed, with 79.67\% of citations falling after 2019, with a median of 2023 and a mode of 2024.} Since dates are not present in ResearchQA citations, this predictor cannot be used for that dataset.
\item[Citation mismatch ($\text{C}_M$)] \textbf{[ResearchQA only]} captures discrepancies between inline citations and the reference list in ResearchQA responses.
Unlike SciArena, which uses author-year inline citations (e.g., ``\texttt{Li} \texttt{et} \texttt{al.} \texttt{2025}''), ResearchQA responses commonly use numbered citations (e.g., ``\texttt{[1]}'', ``\texttt{[2]}'') accompanied by a reference list at the end of the response.
In a minority of cases, inline citations and reference list entries do not match: either an inline citation has no entry in the reference list or vice-versa.
To account for the possibility that judges may penalize such discrepancies, we count the number of mismatched citations $\text{C}_M$.
\end{description}

% Collectively, these variables enable us to assess the relation between several key citation-related features of responses and preference judgments.

To give meaningful interpretations to these citation-related effects, which are often intertwined with response content, we control for two other potential confounds apart from presentation order:
\begin{description}[itemsep=0pt,parsep=1pt]
    \item[Response length ($\text{R}_L,$] in characters\textbf{)}---a common source of bias in preference data~\cite{dubois2024lengthcontrolled, park-etal-2024-disentangling, hu-etal-2025-explaining}.
    \item[Response content ($\text{R}_C$)] measures the difference in unique content between two responses (after removing inline citations). We use BERTScore~\citep{bert-score} precision to capture semantic overlap: the precision of BERTScore(A, B) reflects how much of Response A's content is covered by Response B. We define the proportion of unique content in A relative to B as: $1 - \text{Precision}(A, B)$, where a higher value indicates more unique content.\footnote{We use \texttt{roberta-large} for BERTScore.}
\end{description}

% The second is response content: given two responses with identical citations---one that correctly answers the query and one that doesn't---we expect the former to be preferred, and a control on content allows us to capture this. Thus, we compute \textbf{response similarity ($\text{R}_S$)} measured as the embedding cosine similarity between $R_A$ and $R_B$ with citations redacted (see~\autoref{app:llm-exps}).

We express each continuous predictor as the z-scored difference between Response A ($R_A$) and Response B ($R_B$) for that variable.
No fixed-effects predictors show multicollinearity, with variance inflation factors (VIFs) $< 2$.

% Finally, we consider several citation-related variables, which we express as differences between $R_A$ and $R_B$. One clearly important factor is the total number of citations each response contains: at least up to a point, it is plausible that responses with more citations will be viewed as more credible. A second key factor is the publication date of the cited works. AI papers, for example, have a known bias toward more recent citations~\citep{fang-etal-2025-llm-recency}, and researchers more generally may be inclined to assign greater weight to responses that reference the latest work on a given topic. As such, we compute the mean publication date over all citations in $R_A$ and in $R_B$. Lastly, ...

% \footnote{This intercepts capture the average win-rate $\Delta$ (in log-odds space) between $M_A$ and $M_B$ beyond...}

\paragraph{Random-Effects Design}
\label{setup:random}

Given the differences between the two datasets (e.g., the number of models involved in generating responses), we specify a maximal random-effects structure for each dataset separately~\citep{BARR2013255}. 
We then simplify each maximal model by iteratively eliminating random effects via AIC-based~\cite{aic} model selection to ensure parsimony and minimize overfitting~\cite{MATUSCHEK2017305}, repeating this process for each set of human or LLM judgments.

% \footnote{
% % Eliminated random effects are replaced with fixed effects.
% Fixed effects are unchanged throughout this process;
% \autoref{app:model-selection} has further details on model design and selection.}

\paragraph{Average Marginal Effects} 
We report population-level average marginal effects~\cite{ame, marginaleffects}.
Results are shown as $\Delta \text{Win Rate}$ in percentage points (\textit{pp}): the change in probability that $R_A$ is chosen per 1-SD increase in the predictor, marginalizing over random effects.
Higher values indicate (for each predictor) that $R_A$ has (a) higher citation diversity, (b) more total citations, (c) overall older citations (SciArena) or more mismatched citations (ResearchQA), (d) a longer response, or (e) more unique content.

\section{Case Study: SciArena}
\label{sec:results-sciarena}

For both SciArena (\S\ref{sec:results-sciarena}) and ResearchQA (\S\ref{sec:results-researchqa}) we first discuss details of the data, how those details inform our mixed-effects model design, and what research questions we seek to answer with these models. We then discuss observations about the influence of citation-related features on human preferences and how this compares to the influence of \emph{non-}citation-related features. Finally, we present observations on the four LLM judges.

\subsection{Setup Details}

\paragraph{Data}
SciArena~\cite{zhao2025sciarenaopenevaluationplatform} contains majority preferences from 102 researchers on pairs of LLM-generated responses to scientific queries ($Q$).
We denote an ordered pair of responses as $R_{AB}{=}(R_A, R_B)$ and the ordered pair of models that generated $R_{AB}$ as $M_{AB}{=}(M_A, M_B)$. 
$M_A$ and $M_B$ generate responses given the same set of retrieved documents.
We focus on examples where (i) both $R_A$ and $R_B$ contain $\geq 1$ citation(s) and (ii) the preference label is \emph{either} $R_A$ or $R_B$, filtering cases where $R_A$ and $R_B$ tie or are both deemed bad.
This results in $12,249$ examples, covering $23$ models and $480$ unique model pairs. 
Each $Q$ is labeled with one of six question types ($Q_\text{type}$) and one of $25$ subjects ($Q_\text{subj}$) drawn from one of the five main categories ($Q_\text{cat}$).

\paragraph{Mixed-Effects Model Design}
The rich features of SciArena---including many responses per model ($M_A$/$M_B$), question category ($Q_\text{cat}$), and question type ($Q_\text{type}$)---motivate our use of a relatively complex maximal random effects structure. 
% SciArena contains many responses per model ($M_A$/$M_B$), question category ($Q_\text{cat}$), and question type ($Q_\text{type}$), which motivates our use of random (i.e.\ group-specific) effects. 
We include random \emph{intercepts} for $M_A$, $M_B$, $M_{AB}$, $Q_\text{type}$, and $Q_\text{subj}$, and for interactions defined by $\{M_A,M_B\} \times \{Q_\text{type}, Q_\text{cat}\}$. Second, we include random \emph{slopes} for all continuous variables ($\text{C}_D$, $\text{C}_C$, $\text{C}_A$, $\text{R}_L$, $\text{R}_C$) separately for each $M_{AB}$, $Q_\text{subj}$, and $Q_\text{type}$.\footnote{
This random-effects design, written below (in \href{https://paulrjohnson.net/blog/2022-11-01-multilevel-model-r-cheatsheet/}{\texttt{lme4} syntax}), in the maximal structure:\vspace{-2mm}
\begin{equation*}
\begin{aligned}
    &~~~~~(1|M_A)+(1|M_B) \\
    &+(1|M_A:Q_{\text{type}})+(1|M_B:Q_{\text{type}}) \\
    &+(1|M_A:Q_{\text{cat}})+(1|M_B:Q_{\text{cat}}) \\
    &+(\Delta\text{C}_{D}+\Delta\text{C}_{C}+\Delta\text{C}_{A}+\Delta\text{R}_{L}+\Delta\text{R}_{C}|M_{AB}) \\
    &+(\Delta\text{C}_{D}+\Delta\text{C}_{C}+\Delta\text{C}_{A}+\Delta\text{R}_{L}+\Delta\text{R}_{C}|Q_{\text{subj}}) \\
    &+(\Delta\text{C}_{D}+\Delta\text{C}_{C}+\Delta\text{C}_{A}+\Delta\text{R}_{L}+\Delta\text{R}_{C}||Q_{\text{type}})
\end{aligned}
\end{equation*}
} The continuous fixed-effects predictors are citation diversity, citation count, citation age, response length, and response content. \autoref{app:model-selection} has details on model selection and final model fits.

\paragraph{Research Questions}
For SciArena, we ask:
\begin{enumerate}[leftmargin=*, itemsep=2pt, parsep=-1pt, topsep=0pt]
\item For both human and LLM judges, are more diverse citations preferred or is it only the total citation count that matters? How do the effect sizes for these predictors compare to those for non-citation-related predictors?
\item Do citations' publication dates correlate with preferences, as newer information may reflect more current science than older information?
\end{enumerate}
% We discuss the findings around these hypotheses below.

% We begin by discussing the best model for human judgments (\S\ref{res:human}) then the best models for LLM judgments (\S\ref{res:llm-judge}), concluding with some comparative analysis of their random effects (\S\ref{res:random-effects}).

\begin{figure}[t!]
\centering
\includegraphics[width=\linewidth]{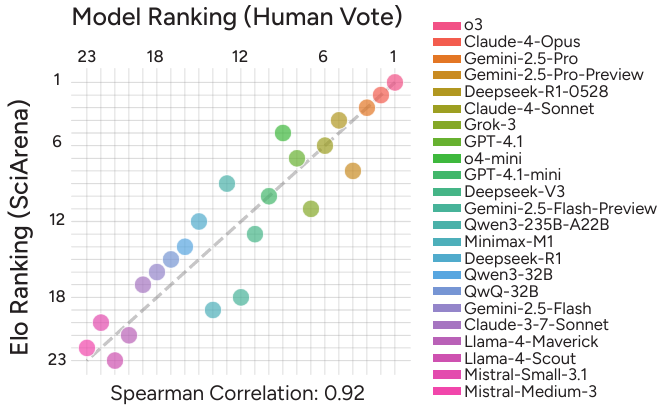}
\caption{Model ranking correlation between Elo Ranking (SciArena) and Random-Effects Ranking (Human).}
\label{fig:human-elo-new}
\end{figure}

\begin{figure}
\centering
\begin{subfigure}{\linewidth}
    \centering
    \caption{\textsc{Citation Diversity} Difference}
    \label{fig:fixed-effects-sciarena-a}
    \includegraphics[width=\linewidth]{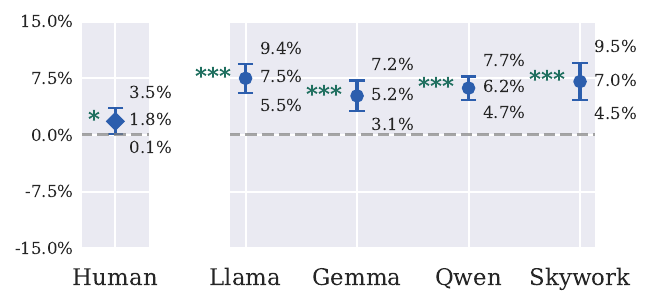}
\end{subfigure}

\begin{subfigure}{\linewidth}
    \centering
    \caption{\textsc{Citation Count} Difference}
    \label{fig:fixed-effects-sciarena-b}
    \includegraphics[width=\linewidth]{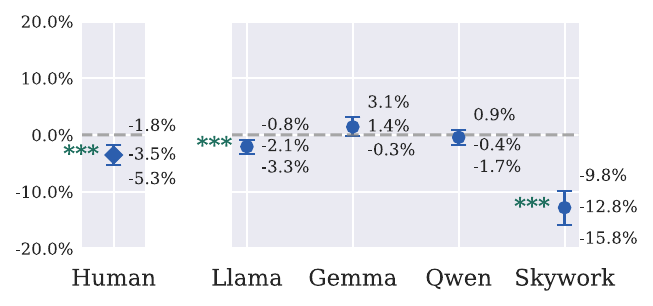}
\end{subfigure}

\begin{subfigure}{\linewidth}
    \centering
    \caption{\textsc{Citation Age} Difference}
    \label{fig:fixed-effects-sciarena-c}
    \includegraphics[width=\linewidth]{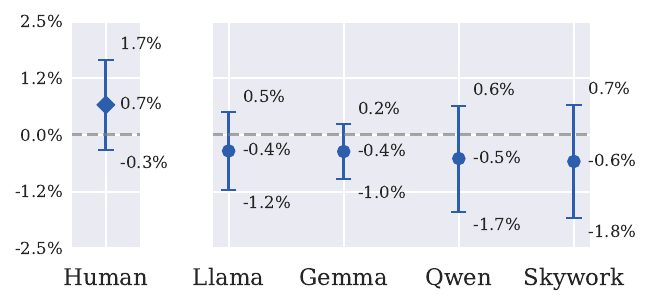}
\end{subfigure}

\begin{subfigure}{\linewidth}
    \centering
    \caption{\textsc{Response Length} Difference}
    \label{fig:fixed-effects-sciarena-d}
    \includegraphics[width=\linewidth]{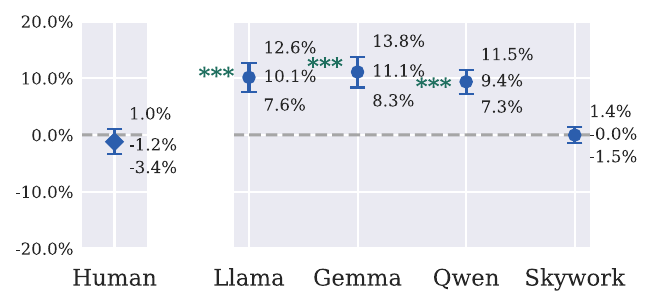}
\end{subfigure}

\begin{subfigure}{\linewidth}
    \centering
    \caption{\textsc{Response Content} Difference}
    \label{fig:fixed-effects-sciarena-e}
    \includegraphics[width=\linewidth]{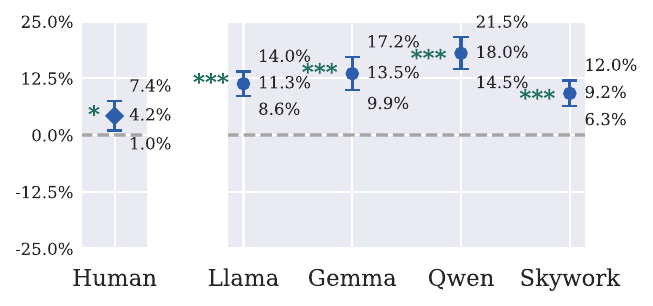}
\end{subfigure}
\caption{\textbf{SciArena}: Estimated fixed effects on human votes and LLM judgments ($\Delta \text{Win Rate}$).
Sub-figures (a)--(c) show citation-related predictors, while (d) and (e) show non-citation control variables for effect size comparison.
Error bars represent 95\% CIs. Significance: ***$p < 0.001$; **$p < 0.01$; *$p < 0.05$.}
\label{fig:fixed-effects-sciarena}
\end{figure}

\subsection{Human Preferences}
\label{res:human-sciarena}
\autoref{fig:fixed-effects-sciarena} plots average marginal effects on win rate ($\Delta$ Win Rate) for the three citation-related predictors and two non-citation variables.\footnote{Exact estimates are reported in Table~\ref{tab:sciarena-estimates} (\autoref{app:model-selection}).} Human preferences are presented on the left.
Here, we observe that a 1$\sigma$ increase in citation diversity difference (i.e.\ when $R_A$ has more diverse citations) yields a $1.8$\% absolute higher probability that $R_A$ is preferred by human judges (\autoref{fig:fixed-effects-sciarena-a}). Conversely, a 1$\sigma$ increase in total citation difference yields a 3.5\% \emph{lower} probability of $R_A$ being selected (\autoref{fig:fixed-effects-sciarena-b}). Thus, human judges prefer \emph{more diverse} citations, but fewer of them \emph{overall}.
The effect size is similar to the response length effect (-1.2 \textit{pp}, n.s.; \autoref{fig:fixed-effects-sciarena-d}) and the preference for semantically unique response content (4.2 \textit{pp}, $p<0.05$; \autoref{fig:fixed-effects-sciarena-e}). We also find that humans show slight preferences for older citations, but the result is not significant (\autoref{fig:fixed-effects-sciarena-c}).

% Further, by calculating a leaderboard ranking based on both $M_A$ and $M_B$ random intercepts, we find a Spearman correlation of 0.98 with the Elo Ranking reported in SciArena (see \autoref{app:full-results}), further justifying our mixed-effects model design.
% Note(hope): given the original ranking doesn't control different factors, does it mean that human are not biased here?

Further, by calculating a leaderboard ranking based on both $M_A$ and $M_B$ random intercepts, we find a Spearman correlation of 0.92 with the Elo Ranking reported in SciArena (see~Figure~\ref{fig:human-elo-new}), further justifying our mixed-effects model design.

\subsection{LLM Judge Preferences}
\label{res:llm-judge-sciarena}
% \paragraph{Citations} 
Compared to humans, LLM judges show similar but stronger preferences for diverse citations ($\Delta$ Win Rate=\{7.5\%, 5.2\%, 6.2\%, 7.0\%\} for Llama, Gemma, Qwen, and Skywork; \autoref{fig:fixed-effects-sciarena-a}).
For total citation count, we observe more mixed results (\autoref{fig:fixed-effects-sciarena-b}). Llama yields a 2.1 \textit{pp} lower probability of $R_A$ being selected, and Skywork presents an even stronger negative correlation (-12.8 \textit{pp}, $p<0.001$). Results for the other two LLM judges are not significant.
None of the LLM judges shows significant citation age-related preferences, though the estimated effects differ in polarity from those of human judges, showing a slight preference for newer citations (\autoref{fig:fixed-effects-sciarena-c}).

For non-citation-related variables, Llama, Gemma, and Qwen all exhibit strong preferences for longer responses (\autoref{fig:fixed-effects-sciarena-d}). By contrast, Skywork's preference is negligible, possibly because it is the only trained reward model among the four.
The response content predictor (which captures response content uniqueness) shows much stronger correlations than those of human judges across all LLM judges (\autoref{fig:fixed-effects-sciarena-e}). While citations can plainly influence how humans perceive information---whether by introducing extra cognitive load or by affecting perceived trustworthiness---LLMs process this information in a fundamentally different manner.
Citations constitute only a small portion of the response, and LLM judges lack the agency to search and verify the cited content (at least in the annotation protocols for SciArena and ResearchQA).
Thus, it is intuitive that differences in response content are more likely to drive LLM predictions.
Controlling for content-related variables thus provides greater confidence in understanding the role of citations in the preferences of LLM judges.\footnote{We further test prompt sensitivity using two variants: (i) a chain-of-thought (CoT) variant that asks the model to reason step by step before deciding, and (ii) a length-debiasing variant that adds an explicit instruction sentence.
Decisions largely remain consistent across variants for all four judges (Cohen's $\kappa \geq 0.67$, $\geq 84\%$ raw agreement under CoT; $\kappa \geq 0.93$, $\geq 97\%$ under length debiasing).
More details are in \autoref{app:llm-exps}.}

\section{Case Study: ResearchQA}
\label{sec:results-researchqa}

% We present the setup details, human preference results, and LLM judge preference results for ResearchQA as follows.

\subsection{Setup Details}

\paragraph{Data} ResearchQA~\citep{yifei2025researchqaevaluatingscholarlyquestion} is a large resource for evaluating LLM systems through curated queries and rubric items based on existing survey articles.
However, only 169 of these examples feature human annotations (from a pool of 31 PhD annotators) that satisfy our criterion of having a clear preference for one response over the other.
These examples span eight scientific fields (e.g., NLP) from five general domains (e.g., Engineering), and all compare responses from the same pair of models, \texttt{gpt-4.1-mini} and \texttt{gemini-2.5-flash}.

%While rubrics are available with these queries, no verified rubric-rating annotations are available.
%We do \emph{not} use LLMs to obtain these rubric-related features, to avoid circularity issues when interpreting LLM judge preferences later on.

\paragraph{Mixed-Effects Model Design}
Given the limited data, we design the maximal random effects structure simply as $(1|Q_\text{field})$ to ensure reasonably robust parameter estimates. For the model trained on human preferences, we randomize the order of responses within each pair so that Response A has an equal chance of being the \texttt{gpt-4.1-mini} or the \texttt{gemini-2.5-flash} response. For the model trained on LLM preferences, we create two versions of each example for both orderings of responses (as in SciArena), and assess position bias through both the (global) fixed effect intercept and a separate presentation order predictor. Since all examples compare the same two models, we do not include by-model random intercepts. 

As noted in \S\ref{sec:setup}, citations in ResearchQA lack dates and thus we eliminate the citation age fixed effect predictor. Additionally, since ResearchQA responses track citations both \emph{inline} and as a reference list at the \emph{end} of the text, we introduce a \emph{citation mismatch} predictor to account for occasional discrepancies between these two lists that may impact preference judgments. See Appendix~\ref{app:model-selection} for details on the final model fits.
% \footnote{If a singular fit warning arises, the Generalized Linear Mixed Model reduces to a Generalized Linear Model.}
% Because all examples compare the same two models, rather than controlling the group variable through random intercepts as in SciArena, for mixed-effects modeling on human votes, we shuffle the data to ensure that Response A has an equal chance of being from \texttt{gpt-4.1-mini} or \texttt{gemini-2.5-flash}.
% For mixed-effects modeling of LLM judge predictions, as we have run inference twice on the same query with both response orders, we address the confound through the presentation-order variable and the fixed intercept.

\paragraph{Research Questions}
% We replace the citation age fixed effect predictor (used in SciArena) with a citation mismatch predictor. As discussed, due to the format difference, citation year information is not present in ResearchQA responses. However, the reference list is more visible under this setup, making citation mismatch a unique feature.
For ResearchQA, we ask:
\begin{enumerate}[leftmargin=*, itemsep=2pt, parsep=-1pt, topsep=0pt]
\item For both human or LLM judges, what is the influence of citation diversity and total citation count on preference judgments?
\item How much does the inconsistency between inline citations and the reference list matter when choosing between responses?
\end{enumerate}

\begin{figure}
\centering
\begin{subfigure}{\linewidth}
    \centering
    \caption{\textsc{Citation Diversity} Difference}
    \label{fig:fixed-effects-researchqa-a}
    \includegraphics[width=\linewidth]{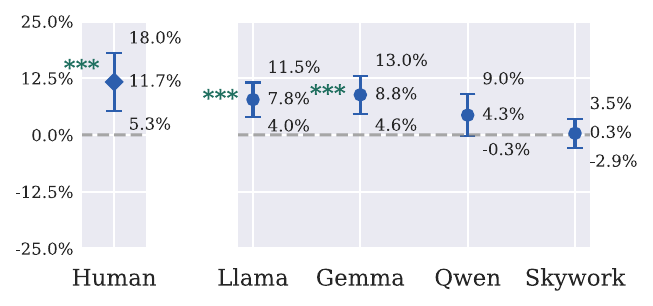}
\end{subfigure}

\begin{subfigure}{\linewidth}
    \centering
    \caption{\textsc{Citation Count} Difference}
    \label{fig:fixed-effects-researchqa-b}
    \includegraphics[width=\linewidth]{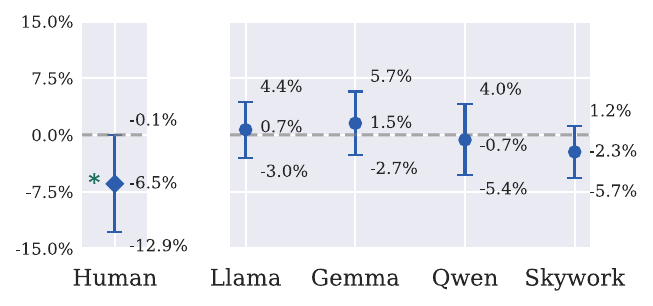}
\end{subfigure}

\begin{subfigure}{\linewidth}
    \centering
    \caption{\textsc{Citation Mismatch} Difference}
    \label{fig:fixed-effects-researchqa-c}
    \includegraphics[width=\linewidth]{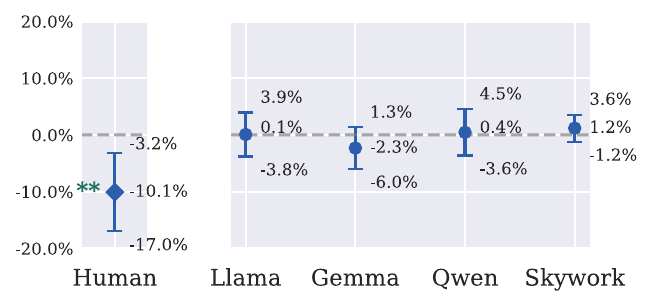}
\end{subfigure}

\begin{subfigure}{\linewidth}
    \centering
    \caption{\textsc{Response Length} Difference}
    \label{fig:fixed-effects-researchqa-d}
    \includegraphics[width=\linewidth]{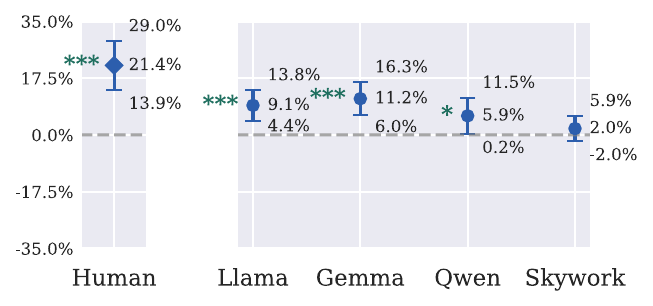}
\end{subfigure}

\begin{subfigure}{\linewidth}
    \centering
    \caption{\textsc{Response Content} Difference}
    \label{fig:fixed-effects-researchqa-e}
    \includegraphics[width=\linewidth]{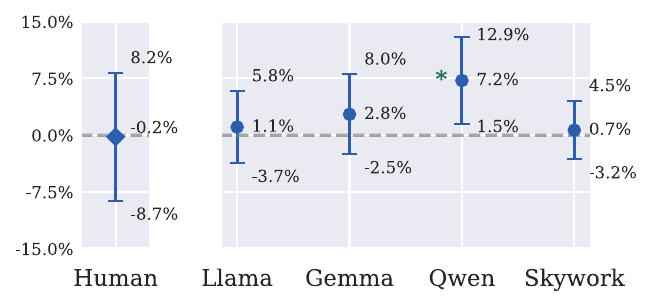}
\end{subfigure}
\caption{\textbf{ResearchQA}: Estimated fixed effects on human votes and LLM judgments ($\Delta \text{Win Rate}$).
Sub-figures (a)--(c) show citation-related predictors, while (d) and (e) show non-citation control variables for effect size comparison.
Error bars represent 95\% CIs. Significance: ***$p < 0.001$; **$p < 0.01$; *$p < 0.05$.}
\label{fig:fixed-effects-researchqa}
\end{figure}

\subsection{Human Preferences}
\autoref{fig:fixed-effects-researchqa} shows the observed effects.\footnote{Exact estimates are reported in Table~\ref{tab:researchqa-estimates} (\autoref{app:model-selection}).}
% Though the preference collection processes differ (i.e., a different group of annotators and a different annotation interface),
Despite the differences in the preference data collection process between ResarchQA and SciArena, we obtain consistent results: here again, human judges prefer more diverse citations but fewer total citations (\autoref{fig:fixed-effects-researchqa-a}, \autoref{fig:fixed-effects-researchqa-b}). Furthermore, the citation mismatch predictor significantly negatively correlates with a response's probability of being chosen, suggesting that humans penalize discrepancies between inline citations and the reference list (e.g., due to hallucinated citations)---a sensible reaction (\autoref{fig:fixed-effects-researchqa-c}).
The effect sizes are fairly large, with $\Delta$ Win Rate = \{11.7\%, -6.5\%, -10.1\%\} for citation diversity, citation count, and citation mismatch. By comparison, $\Delta$ Win Rate = 21.4\% for response length difference, while no significant effects are observed for response content difference as modeled here (\autoref{fig:fixed-effects-researchqa-d}, \autoref{fig:fixed-effects-researchqa-e}).\footnote{Note that the observed human preferences on response length and content differ from SciArena, likely due to differences in annotator populations, collection setups, and response comparisons. We leave further investigation to future work.}
% \ww{It may sound a bit odd to say that we find ``no effect of response content.'' This might benefit from some elaboration.}

\subsection{LLM Judge Preferences}
For citation-related predictors (\autoref{fig:fixed-effects-researchqa-a}, \autoref{fig:fixed-effects-researchqa-b}, \autoref{fig:fixed-effects-researchqa-c}), we find significant associations only between citation diversity and preferences from Llama and Gemma, with effect sizes of 7.8 \textit{pp} and 8.8 \textit{pp}, respectively. Citation diversity effects are also positive for Qwen and Skywork, though not significant.
For citation count and citation mismatch, we do \emph{not} observe the same strongly negative effects that we saw for human judges, revealing an interesting divergence in the citation-related preferences between the two judge categories.
% This may further highlight the gap between humans and LLMs, and the downstream decision biases arising from these differences (e.g., mismatched citations signal poor responses for human judges, yet LLM judges do not penalize them).

Consistent with our SciArena results, Llama, Gemma, and Qwen all show significant bias toward longer responses ($\Delta$ Win Rate = \{9.1\%, 11.2\%, 5.9\%\}), while Skywork (a trained reward model) does not (\autoref{fig:fixed-effects-researchqa-d}).
For response content difference (\autoref{fig:fixed-effects-researchqa-e}), only Qwen shows a preference for unique content (7.2 \textit{pp}, $p<0.05$).

\paragraph{Sensitivity to Citation Perturbations}
To probe how citation \emph{validity} and surface cues affect judge preferences, we run four citation-perturbation experiments on ResearchQA.
For each item and condition, we randomly select one of the two responses and perturb its reference list, leaving the response text and inline \texttt{[N]} markers unchanged:
(1) replace each publication year with a future year,
(2) replace each year with a random incorrect past year,
(3) shuffle the words within each title,
or (4) permute URLs across the reference list, breaking the original title--URL pairings.
% \footnote{The last condition requires at least two references on the perturbed side and covers 126 of the 169 items; the other three cover all 169 ($\times$ 2 presentation orders).}
% Unlike the other conditions, URL permutation preserves both title wording and publication years.
We run four judges 
% without access to the cited sources 
and report $\Delta$, the decrease in preference for the perturbed response relative to its intact-citation baseline.\footnote{The last condition requires at least two references on the perturbed side and covers 126 of the 169 items; the other three cover all 169 ($\times$ 2 presentation orders).
We average preferences across the two presentation orders within each item and use a paired Wilcoxon signed-rank test. Positive $\Delta$ indicates reduced preference after perturbation.}

All four judges show reduced preference for the perturbed response under the first three conditions ($\Delta = +5.9$ to $+20.1$ percentage points, all $p<0.001$).
Title--URL permutation produces smaller shifts ($+2.0$ to $+4.0$ points), significant only for Gemma and Skywork.
Random incorrect years produce larger decreases than future years for every judge (by $2.7$--$9.2$ points), despite future dates being overtly implausible.
Among items where the selected response is preferred in both baseline presentation orders, title shuffling causes the judge to choose the alternative in at least one order for $36$--$46\%$ of items; random incorrect years do so for $37$--$67\%$, depending on the judge.
These results demonstrate sensitivity to citation perturbations, but leave open the extent to which LLM judges base their preferences on citation validity.

\section{Discussion}
\label{sec:discussion}

% We summarize and discuss the implications of our findings from SciArena and ResearchQA.

\subsection{Collecting Human Preference Data}
\label{subsec:collecting-human-preferences}
% In both \S\ref{sec:results-sciarena} and \S\ref{sec:results-researchqa}, we found that human judges prefer responses with more diverse citations but fewer overall. Furthermore, depending on the task, response content or response length also exhibit significant effects on human preferences. Finally, for ResearchQA, we additionally noted a strong \emph{negative} correlation between citation inconsistency and the probability of a response being chosen. \ww{Depending on how much content we add back in the main text, this initial summary paragraph can possibly be omitted.}
% This motivates us to think further about the role of citations in human preference judgments.
A key fact about many preference data annotation protocols is that they tend to \emph{underspecify} the criteria by which responses ought to be judged. Especially when annotators are domain experts, it is often viewed as beyond the purview of those collecting the data to prescribe a set of factors to be considered in making such judgments. Rather, these factors are to be learned \emph{implicitly} by a reward model later trained on those judgments.

Yet, even experts are not immune from lazy or heuristic assessments of citation quality---a tendency that is especially undesirable in (ostensibly rigorous) academic settings.  This reflects a broader pattern: information source preferences are influenced by internal factors such as user's cognition and affection~\cite{10.1108/JD-03-2023-0059, 10.1145/3673791.3698409}. 
% \ww{Might be nice if we could find a citation for this}. 
As such, those wanting to discourage these practices during data collection may find it valuable to provide clear citation-related rubrics or tools that assist annotators in more easily understanding cited content (e.g., reading assistants that surface key source passages~\cite{10.1145/3729176.3729196}). Furthermore, the effects of particular protocol design choices should be at least partially observable with MEMs, giving data consumers (e.g., post-training or evaluation teams) confidence that the resulting preference data reflects desired behavior.

% First, we note that preference data collection protocols do not typically specify how citations \emph{should} factor into judges' assessments, risking inconsistent or subjective evaluation of citations.
% This is especially undesirable for academic applications.
% Meanwhile, judges with expertise in the target domain likely have internalized its citation-related norms, but others (even experts in adjacent fields) likely have not. In the latter case, it is valuable to provide clear evaluation rubrics and tools that assist judges in understanding cited content (e.g., reading assistants that motivate judges to check details critically rather than only skimming titles).

\subsection{Collecting LLM Preference Data}
\label{subsec:collecting-llm-preferences}
Recruiting human judges to produce preferences on long-form responses can be both costly and challenging. While it is much easier to set up an LLM judge to approximate human preferences, our analysis reveals that LLM judges exhibit substantially different preference patterns from humans. Moreover, these patterns are data-dependent: in some cases, citation-related factors seem to exhibit similar influences on both human and LLM judges, while in others, they may exhibit undesirable biases on superficial features (e.g., citation year, response length) or diverge from humans in other respects (e.g., in how citation inconsistencies are penalized).

One straightforward way to reduce these divergences is to ensure that LLM judges have access to cited documents through external tools~\citep{findeis-etal-2025-external}---just as human judges may search a cited source on the web or preview them in the annotation interface itself. Additionally, for domains with clearly established best practices for citations, domain-specific rubrics (of the kind discussed above) may be employed to ensure that LLM judges consider the extent to which a response emulates those standards~\citep{farzi-dietz-2025-criteria}. Such factors demand careful consideration when designing LLM judge systems.

\subsection{Choosing an LLM Judge}
\label{subsec:choosing-an-llm-judge}
Finally, we consider the implications of the differences observed among LLM judges.

First, it is evident that LLM judges \emph{do} exhibit diverse preferences. For instance, models show quite different degrees of preference for unique response content. In some cases, these differences may indicate substantial differences in training---as in the lack of response length bias in Skywork compared to the other three models.

Second, and as a consequence of this first observation, assessing LLM judge quality simply via accuracy or correlation relative to human preferences risks obscuring idiosyncrasies of individual judge models. Analysis with MEMs of the sort conducted here is one tool that can aid in measuring biases of interest and in facilitating more nuanced and principled LLM judge selection. When collecting preference data, it is thus imperative to consider not only \emph{whether} to use an LLM judge, but also \emph{which} LLM judge to use and what biases are or are not acceptable for one's use case.

% When treating human preferences as labels, we can calculate standard task metrics (e.g., accuracy) for each LLM judge.
% Despite the differences in accuracy usually being quite small, we show that the four LLM judges tested have different preferences.
% This points out the need to interpret preferences rather than only understanding model behaviors through aggregated scores.
% In fact, our mixed-effects models are able to reflect, to some extent, how a model is trained or instructed to behave. For example, response length bias has been widely studied in preference data, but the trained reward model Skywork shows less of this bias compared to the other three LLM judges.

% Therefore, when collecting preference data annotations, it is important to consider not only whether to use an LLM judge, but also which LLM judge to use and what assumptions or tolerances are acceptable for each preference dimension.

\section{Related Work}

\paragraph{Human/LLM Preferences}
Pairwise comparisons have been widely used to understand human preferences.
Among these efforts, a first line of work focuses on evaluation using human preferences for a variety of complex tasks, such as information-seeking via search and coding~\cite{chatbotarena, miroyan2025searcharenaanalyzingsearchaugmented, chi2025copilot, BIANCHI2026100943}.
A second line of work proposes different approaches to mine or uncover latent preferences from aggregated human pairwise votes~\cite{zhong-etal-2024-explaining, findeis2025InverseConstitutionalAI, movva2025whatshumanfeedbacklearning}.
Further, given the difficulty of collecting high-quality human preference labels, existing works use LLM judges to obtain similar pairwise votes. Despite the greater convenience and scalability of LLM judgments, much of this work has also discussed potential LLM or LLM judge biases~\cite{bharadwaj2025flatteryflufffogdiagnosing, shi-etal-2025-judging, tan2025judgebench, gu2026survey, dunlap_vibecheck} and how to better align LLM judges to humans~\cite{chen-etal-2024-humans, liu2024aligning}.
While a few works investigate citation-related preferences---focusing either on biases in how LLMs use citations compared to humans~\cite{ando2026aligninglargelanguagemodel, algaba-etal-2025-large} or on building systems to embed citations in long-form responses~\cite{huang-etal-2024-training, zhang-etal-2025-longcite}---to the best of our knowledge, no prior work has attempted to explain the role of citations in preference data for both human and LLM judges.

\paragraph{Deep Research Applications}
While our analysis focuses on understanding preferences to better inform preference data collection, citations are also central to deep research applications, where complex systems produce long-form responses with citations for challenging tasks~\cite{singh-etal-2025-ai2, shao2026dr}.
Many deep research benchmarks have been proposed, including DeepResearch Bench~\cite{du2026deepresearch}, ResearchRubrics~\cite{sharma2026researchrubrics}, SurGE (Survey Generation Evaluation)~\cite{su2026surgebenchmarkevaluationframework}, and AstaBench~\cite{bragg2025astabenchrigorousbenchmarkingai}, among others.
\citet{hwang2026deepresearchshallowevaluation} study the challenges in deep research system evaluation and provide practical suggestions for designing future meta-evaluations.
Survey generation and literature review are prominent applications of deep research, presenting both significant opportunities and challenges~\cite{agarwal2025litllms, yang2026deepsurveyenhancinganalyticaldepth, sahu2026rethinkingliteraturesearchevaluation}.
Understanding citation preferences and the underlying human preferences or LLM behaviors is therefore essential for enabling more meaningful evaluation of these systems.

\section{Conclusion}
\label{sec:conclusion}

This work has studied how human and LLM judges weigh citations in their assessment of response quality, controlling for other potential confounds. By studying human and LLM preferences on the same data, we are able to compare predictors' effects both between these two categories of judge and across different types of LLM judge.
% We find evidence of rational human behavior regarding citations: diverse citations are preferred, while possibly redundant or erroneous citations are not.
% LLM judges show mixed preferences compared to humans, indicating different mechanisms at play when performing the pairwise preference annotation task.

Broadly, we find that human judges prefer diverse citations but fewer total citations, while LLM judges exhibit different citation-related preferences depending on the data and model.
We observe cases where all LLM judges exhibit an identifiable bias (e.g., response length on SciArena), while humans show no reliable preference along the same dimension. Further, we observe cases where humans show certain rational behaviors (e.g., penalizing mismatched citations on ResearchQA) that LLM judges do not.
For LLM judges, different training processes also seem to engender different preferences for certain features (e.g., response length).

Taken together, these observations contribute to our understanding of the role of citations in preference data from human and LLM judges. Based on these observatins, we also provide some suggestions for preference data collection: both human and LLM judges may benefit from more detailed instructions for assessing citation quality before and during the preference collection process---with potential aids including clear annotation rubrics and external tool access. Mixed-effects models can also be integrated into the pipeline to help ensure data quality. More generally, we hope this work will motivate more careful attention to cited sources throughout the preference data collection pipeline.

\section*{Limitations}

\paragraph{Correlation vs.\ Causality}
We present results from the final mixed-effects models and emphasize that these results reflect correlations, not causal relationships.
We do not claim that these features \emph{causally} affect human or LLM preference. Importantly, however, features need not be causal in order for LLM judges to learn to rely on them.

\paragraph{Dataset.} We focus exclusively on SciArena and ResearchQA to study human and LLM preferences, as these are the only two datasets we could find that satisfied our research needs (\S\ref{sec:setup}). As~\citet{movva2025whatshumanfeedbacklearning} note, different datasets can yield different observations about human preferences, often depending on the specific queries used and annotators involved.
However, we believe that our mixed-effects modeling approach is widely applicable and that, given the broad coverage of the datasets, many of our findings likely generalize. 

\paragraph{Prompting.} While we adopt the same prompt setup as SciArena, different prompts could lead to variations in observed model preferences. We test sensitivity to prompt variation in \autoref{app:llm-exps}, where results suggest that LLM judge decisions largely remain unchanged. Substantially different prompts could still produce different behavior, which we leave to future work.
% Specifying a distinctive persona in the system prompt, such as an \emph{expert in scientific literature synthesis}, could introduce unexpected effects.
% To test the sensitivity of the prompts to length and positional bias specifically, we conducted additional experiments with Qwen3-30B by incorporating instructions such as ``Do NOT allow the length of the responses to influence your evaluation'' and ``Do NOT let the order in which the candidate responses are presented influence your decision,'' following \citet{liu-etal-2025-metafaith}.
% However, the results correlate highly with those obtained under the original prompts, suggesting that such instruction variants have minimal impact on the outcome.

\paragraph{Linearity Assumption.} To enable interpretable analysis of the impact of citations on preferences, we include the fixed-effects predictors assuming linearity.
Given the complex nature of human and AI decision-making, some variables may relate to the outcome in more complex ways---for example, a longer response may be preferred only up to a point, or the effect of citation diversity may depend on response length.
To probe the latter possibility, we re-run the mixed-effects analysis on all final models with an added interaction between response length and citation diversity. Across all 10 final models (1 human and 4 LLM judges on 2 datasets), the interaction is significant only for the SciArena Llama judge.
We encourage future work to explore potential nonlinear or interaction effects.

\section*{Ethics Statement}
Since this work focuses on the interpretability of preference data and relies on existing public, anonymized datasets, we do not believe it raises any substantive ethical concerns.
Both datasets used in our analysis are publicly available and contain no personally identifiable information.
The LLM judges evaluated in this work are open-weight models accessed through HuggingFace, and our analysis is observational---aimed only at understanding citation-related preferences.

\section*{Acknowledgments}

We are grateful to collaborators and visiting researchers at the HLTCOE at Johns Hopkins University for feedback on this work, including Dayeon (Zoey) Ki, Maxime Dassen, Roxana Petcu, Jia-Huei Ju, Siddharth Singh, Hannah Recknor, James Mayfield, Dawn Lawrie, Laura Dietz, Eugene Yang, Daniel Khashabi, and Kevin Duh---with particular thanks to Bryan Li and Gabrielle Kaili-May Liu for especially detailed input and comments. 
We would like to extend our thanks to the anonymous reviewers for their valuable feedback, and to Navita Goyal, Nishant Balepur, and Christabel Acquaye for their suggestions on the draft.

\bibliography{custom}

% \clearpage
\newpage
\appendix
\section{Experiment Details}
\label{app:llm-exps}

\subsection{LLM Judge Experiment Setup}

We set up the zero-shot experiments using the same prompt as in SciArena~\citep{zhao2025sciarenaopenevaluationplatform}. Hugging Face model names are: 
\nolinkurl{meta-llama/Llama-3.3-70B-Instruct}, 
\nolinkurl{google/gemma-3-27b-it}, 
\nolinkurl{Qwen/Qwen3-30B-A3B-Instruct-2507}, and 
\nolinkurl{Skywork/Skywork-Critic-Llama-3.1-70B}.

We serve the LLMs using vLLM~\cite{kwon2023efficientmemorymanagementlarge}, with 4 NVIDIA A100 GPUs for Llama and Skywork and 1 A100 for Gemma and Qwen. The model inference implementation is based on the SciArena public codebase (with temperature 0).\footnote{\url{https://github.com/yale-nlp/SciArena}}

\subsection{Flipped Response Order Experiment}

We include a response presentation order variable to control for potential positional bias in citation preferences. We use the same prompt template for the flipped-order experiment but switch the order of $R_A$ and $R_B$ as shown in the prompt below.
% This design choice is based on our preliminary LLM judge prediction results where we find LLMs tend to overly pick one option (\autoref{tab:confusion-matrix}). In detail, Llama is biased to return \texttt{A}, whereas Gemma, Qwen, and Skywork are biased to return \texttt{B}. This is consistent with our estimated fixed effects on response order in~\autoref{tab:fixed-predictor-response}.

\begin{prompt}[title={LLM Judge Prompt (Flipped Order)}]
\texttt{You are an expert in scientific literature synthesis. Your task is to evaluate the quality of two AI-generated citation-attributed responses to a user's question. Assess both responses for relevance, accuracy, clarity, and appropriate use of citations. Then, select the response, Output (a) or Output (b), that best address the user's question.} \\ \\
\texttt{User Question:} \\
\texttt{\{question\}} \\ \\
\texttt{Output (a):} \\
\texttt{\{response\_b\}} \\ \\
\texttt{Output (b):} \\
\texttt{\{response\_a\}} \\ \\
\texttt{Which is best, Output(a) or Output (b)?} \\
\texttt{Answer with only: "A" or "B"}
\end{prompt}

\subsection{Prompt Sensitivity Analysis}

To assess whether our findings depend on the specific judge prompt, we re-run all LLM judges (Qwen, Gemma, Llama, and Skywork) on SciArena using two variants of the original prompt: a chain-of-thought (CoT) variant, which asks the model to reason step by step before committing to a decision, and a length-debiasing variant (following \citet{liu-etal-2025-metafaith}), which adds an explicit instruction not to let response length influence the evaluation. Both variants are otherwise identical to the original prompt. As in the main experiments, each pair is judged under both presentation orders. We quantify stability as the agreement between each variant's decisions and those of the original prompt, using Cohen's $\kappa$ alongside raw agreement.

\begin{prompt}[title={LLM Judge Prompt (CoT)}]
\texttt{You are an expert in scientific literature synthesis. Your task is to evaluate the quality of two AI-generated citation-attributed responses to a user's question. Assess both responses for relevance, accuracy, clarity, and appropriate use of citations. Then, select the response, Output (a) or Output (b), that best addresses the user's question.} \\ \\
\texttt{User Question:} \\
\texttt{\{question\}} \\ \\
\texttt{Output (a):} \\
\texttt{\{response\_a\}} \\ \\
\texttt{Output (b):} \\
\texttt{\{response\_b\}} \\ \\
\texttt{Which is best, Output (a) or Output (b)?} \\
\texttt{Let's think step by step, then give your final answer.} \\
\texttt{End your response with a line formatted exactly as:} \\
\texttt{Answer: <A or B>}
\end{prompt}

\begin{prompt}[title={LLM Judge Prompt (Length Debiasing)}]
\texttt{You are an expert in scientific literature synthesis. Your task is to evaluate the quality of two AI-generated citation-attributed responses to a user's question. Assess both responses for relevance, accuracy, clarity, and appropriate use of citations. Then, select the response, Output (a) or Output (b), that best addresses the user's question.} \\ \\
\texttt{User Question:} \\
\texttt{\{question\}} \\ \\
\texttt{Output (a):} \\
\texttt{\{response\_a\}} \\ \\
\texttt{Output (b):} \\
\texttt{\{response\_b\}} \\ \\
\texttt{Which is best, Output (a) or Output (b)?} \\
\texttt{Do NOT allow the length of the responses to influence your evaluation.} \\
\texttt{Answer with only: "A" or "B"}
\end{prompt}

Table~\ref{tab:prompt-sensitivity} below reports the results. 
% Agreement is high under both variants: the CoT reformulation introduces more variability ($\kappa$ from 0.67 to 0.89) than the length-debiasing instruction ($\kappa$ from 0.93 to 0.98). Gemma is the least stable judge under CoT and Skywork the most stable under both variants. 
Overall, these results suggest that our judges' decisions are largely stable under prompt variation.

\begin{table}
\centering
\small
\begin{tabular}{lcccc}
\toprule
 & \textbf{Qwen} & \textbf{Gemma} & \textbf{Llama} & \textbf{Skywork} \\
\midrule
CoT & .80 (90\%) & .67 (84\%) & .72 (88\%) & .89 (95\%) \\
LD & .96 (98\%) & .93 (97\%) & .93 (97\%) & .98 (99\%) \\
\bottomrule
\end{tabular}
\caption{Agreement between each prompt variant and the original prompt, reported as Cohen's $\kappa$ with raw agreement in parentheses. LD: Length Debiasing.}
\label{tab:prompt-sensitivity}
\end{table}

\section{MEM Details \& Model Selection}
\label{app:model-selection}

Although there is not yet consensus on how to select the random-effects design matrix, we follow the recommendation to ``\textit{keep it maximal}''. This leverages the advantage of mixed-effects models over fixed-effects models: the ability to isolate complex group-level variance.
We start with the maximal random effects structure feasible given our data groups, removing terms to ensure a non-singular fit, and further simplifying the random structure based on AIC.\footnote{\url{https://www.rdocumentation.org/packages/lme4/versions/1.1-38/topics/isSingular}}
Note that this approach is relatively costly because fitting the complex model requires more computation and likely over-specifies the possible data generation process.

Specifically, data groups in \textbf{SciArena} include: (1) models ($M_A$ and $M_B$;
% shown in~\autoref{fig:human-elo};
$N=23$); (2) query types ($\lvert Q_\text{type} \rvert=6$); (3) query subjects ($\lvert Q_\text{subj}\rvert = 25$, which further belong to 5 main categories, $Q_\text{cat}$); and (4) model pairs ($\lvert M_{AB}\rvert=480$).
The six question types are: Conceptual Explanation, Challenges \& Limitations, State-of-the-Art Assessment, Methodology Inquiry, Paper Finding, and Others. The five main categories include: Engineering, Natural Science, Healthcare, Humanities \& Social Science, and Others. Further, for example, Computer Science, Electrical Engineering, and Civil Engineering are subjects under Engineering category. Readers can check SciArena~\citep{zhao2025sciarenaopenevaluationplatform} for full details.
% Given that we lack annotator information in SciArena, we consider the main categories/subjects as proxies for this information.

When designing \textbf{random intercepts}, we start by considering the number of groups. It is usually suggested to include variables as random effects when the groups are drawn from a larger population with more than 5 levels. (Therefore, during this step, we select $Q_\text{subj}$ rather than $Q_\text{cat}$.)
For \textbf{interactions}, we hypothesize that different models might show different response quality based on the queries. To balance the number of observations available to estimate a group, we define interactions between $\{M_A,M_B\}$ and $\{Q_\text{type}, Q_\text{cat}\}$. For example, using $M_A : Q_\text{subject}$ leads to at most $23 \times 25$ groups, resulting in fewer observations per group, whereas using $Q_\text{cat}$ results in at most $23 \times 5$ groups. As a result, $M:Q_\text{type}$ leads to 138 groups and $M:Q_\text{cat}$ leads to 115 groups.
Furthermore, regarding \textbf{random slopes}, we include all continuous fixed-effects predictors where we suspect possible group-level variance. The maximal form includes intercepts, slopes, and their interactions (denoted with $|$ in \S\ref{sec:setup}). The exception in our case is the $Q_\text{type}$ groups, which have too few levels to estimate the parameters reliably. Therefore, we have to assume there is no interaction between the question type intercept and slopes in the maximal design, using the syntax $||$.
On the fixed-effects side, we retain all predictors during the process of simplifying the random effects, as these are the main effects we hope to observe.
The final mixed-effects models:
\begin{equation*}
\begin{aligned}
    &\text{(1) Human vote} \sim \\
    & \Delta\text{C}_{D}+\Delta\text{C}_{C}+\Delta\text{C}_{A}+\Delta\text{R}_{L}+\Delta\text{R}_{C} \\
    &+(1|M_A)+(1|M_B) \\
    &+(1|M_B:Q_{\text{type}}) \\
    &+(1|M_A:Q_{\text{cat}})+(1|M_B:Q_{\text{cat}}) \\
    &+(\Delta\text{C}_{D}+\Delta\text{C}_{A}+\Delta\text{R}_{L}+\Delta\text{R}_{C}||M_{AB}) \\
    &+(0+\Delta\text{C}_{C}+\Delta\text{C}_{A}+\Delta\text{R}_{L}+\Delta\text{R}_{C}||Q_{\text{subj}}) \\
    &+(\Delta\text{C}_{D}+\Delta\text{R}_{C}||Q_{\text{type}})
\end{aligned}
\end{equation*}
\begin{equation*}
\begin{aligned}
    &\text{(2) Llama vote} \sim \\
    & \Delta\text{C}_{D}+\Delta\text{C}_{C}+\Delta\text{C}_{A}+\Delta\text{R}_{L}+\Delta\text{R}_{C} + \text{order} \\
    &+(1|M_A)+(1|M_B) \\
    &+(1|M_A:Q_{\text{type}})+(1|M_B:Q_{\text{type}}) \\
    &+(1|M_A:Q_{\text{cat}})+(1|M_B:Q_{\text{cat}}) \\
    &+(\Delta\text{C}_{D}+\Delta\text{R}_{L}||M_{AB}) \\
    &+(\Delta\text{C}_{C}+\Delta\text{C}_{D}+\Delta\text{R}_{C}||Q_{\text{subj}}) \\
    &+(\Delta\text{C}_{A}+\Delta\text{R}_{L}||Q_{\text{type}})
\end{aligned}
\end{equation*}
\begin{equation*}
\begin{aligned}
    &\text{(3) Gemma vote} \sim \\
    & \Delta\text{C}_{D}+\Delta\text{C}_{C}+\Delta\text{C}_{A}+\Delta\text{R}_{L}+\Delta\text{R}_{C} + \text{order} \\
    &+(1|M_A)+(1|M_B) \\
    &+(1|M_A:Q_{\text{type}}) \\
    &+(1|M_B:Q_{\text{cat}}) \\
    &+(\Delta\text{C}_{D}+\Delta\text{C}_{C}+\Delta\text{C}_{A}+\Delta\text{R}_{L}||M_{AB}) \\
    &+(\Delta\text{C}_{D}+\Delta\text{C}_{C}+\Delta\text{R}_{L}+\Delta\text{R}_{C}||Q_{\text{subj}}) \\
    &+(\Delta\text{C}_{D}+\Delta\text{C}_{C}+\Delta\text{R}_{L}+\Delta\text{R}_{C}||Q_{\text{type}})
\end{aligned}
\end{equation*}
\begin{equation*}
\begin{aligned}
    &\text{(4) Qwen vote} \sim \\
    & \Delta\text{C}_{D}+\Delta\text{C}_{C}+\Delta\text{C}_{A}+\Delta\text{R}_{L}+\Delta\text{R}_{C} + \text{order} \\
    &+(1|M_A)+(1|M_B) \\
    &+(1|M_A:Q_{\text{type}})+(1|M_B:Q_{\text{type}}) \\
    &+(1|M_A:Q_{\text{cat}})+(1|M_B:Q_{\text{cat}}) \\
    &+(\Delta\text{C}_{D}+\Delta\text{C}_{C}+\Delta\text{R}_{L}+\Delta\text{R}_{C}||M_{AB}) \\
    &+(\Delta\text{C}_{D}+\Delta\text{C}_{A}+\Delta\text{R}_{L}+\Delta\text{R}_{C}||Q_{\text{subj}}) \\
    &+(\Delta\text{C}_{A}+\Delta\text{R}_{L}+\Delta\text{R}_{C}||Q_{\text{type}})
\end{aligned}
\end{equation*}

\begin{equation*}
\begin{aligned}
    &\text{(5) Skywork vote} \sim \\
    & \Delta\text{C}_{D}+\Delta\text{C}_{C}+\Delta\text{C}_{A}+\Delta\text{R}_{L}+\Delta\text{R}_{C} + \text{order} \\
    &+(1|M_A)+(1|M_B) \\
    &+(1|M_A:Q_{\text{type}})+(1|M_B:Q_{\text{type}}) \\
    &+(1|M_A:Q_{\text{cat}})+(1|M_B:Q_{\text{cat}}) \\
    &+(\Delta\text{C}_{C}+\Delta\text{C}_{A}+\Delta\text{R}_{C}||M_{AB}) \\
    &+(\Delta\text{C}_{D}+\Delta\text{C}_{C}+\Delta\text{C}_{A}+\Delta\text{R}_{L}+\Delta\text{R}_{C} \\
    &||Q_{\text{subj}}) \\
    &+(\Delta\text{C}_{D}+\Delta\text{C}_{A}+\Delta\text{R}_{C}||Q_{\text{type}})
\end{aligned}
\end{equation*}

For \textbf{ResearchQA}, because we have less data groups (e.g., responses are only from two models: \texttt{gpt-4.1-mini} and \texttt{gemini-2.5-flash} and only $169$ observations, the maximal form is: $(1|Q_\text{field})$, where they are in total 8 fields. The final mixed-effects model for human preference is:
\begin{equation*}
\begin{aligned}
    &\text{Human vote} \sim \\
    & \Delta\text{C}_{D}+\Delta\text{C}_{C}+\Delta\text{C}_{M}+\Delta\text{R}_{L}+\Delta\text{R}_{C} + (1|Q_\text{field})
\end{aligned}
\end{equation*}
\noindent The final MEM for LLM judge preference is:
\begin{equation*}
\begin{aligned}
    &\text{Llama / Gemma / Qwen / Skywork vote} \sim \\
    & \Delta\text{C}_{D}+\Delta\text{C}_{C}+\Delta\text{C}_{M}+\Delta\text{R}_{L}+\Delta\text{R}_{C} + \text{order}
\end{aligned}
\end{equation*}

The estimated fixed effects of the final model on human votes and LLM judgments are reported in Table~\ref{tab:sciarena-estimates} (SciArena) and Table~\ref{tab:researchqa-estimates} (ResearchQA).

\begin{table*}[t]
\centering
% \small
% \setlength{\tabcolsep}{5pt}
\begin{tabular}{lllrr}
\toprule
Judge & Predictor & $\Delta$Win Rate & 95\% CI & $p$ \\
\midrule
Human & Citation Diversity & +1.78$^{*}$ & [+0.06, +3.51] & 0.043 \\
 & Citation Count & -3.55$^{***}$ & [-5.34, -1.75] & $<$0.001 \\
 & Citation Age & +0.66\phantom{$^{***}$} & [-0.33, +1.66] & 0.193 \\
 & Response Length & -1.21\phantom{$^{***}$} & [-3.40, +0.99] & 0.283 \\
 & Response Content & +4.21$^{*}$ & [+1.00, +7.43] & 0.010 \\
\midrule
Llama & Citation Diversity & +7.47$^{***}$ & [+5.55, +9.39] & $<$0.001 \\
 & Citation Count & -2.07$^{***}$ & [-3.29, -0.85] & $<$0.001 \\
 & Citation Age & -0.35\phantom{$^{***}$} & [-1.21, +0.50] & 0.417 \\
 & Response Length & +10.13$^{***}$ & [+7.61, +12.64] & $<$0.001 \\
 & Response Content & +11.30$^{***}$ & [+8.61, +13.98] & $<$0.001 \\
\midrule
Gemma & Citation Diversity & +5.15$^{***}$ & [+3.12, +7.19] & $<$0.001 \\
 & Citation Count & +1.41\phantom{$^{***}$} & [-0.27, +3.09] & 0.100 \\
 & Citation Age & -0.37\phantom{$^{***}$} & [-0.98, +0.24] & 0.237 \\
 & Response Length & +11.07$^{***}$ & [+8.35, +13.80] & $<$0.001 \\
 & Response Content & +13.52$^{***}$ & [+9.89, +17.15] & $<$0.001 \\
\midrule
Qwen & Citation Diversity & +6.18$^{***}$ & [+4.65, +7.71] & $<$0.001 \\
 & Citation Count & -0.44\phantom{$^{***}$} & [-1.73, +0.86] & 0.510 \\
 & Citation Age & -0.52\phantom{$^{***}$} & [-1.69, +0.64] & 0.380 \\
 & Response Length & +9.35$^{***}$ & [+7.26, +11.45] & $<$0.001 \\
 & Response Content & +18.00$^{***}$ & [+14.50, +21.50] & $<$0.001 \\
\midrule
Skywork & Citation Diversity & +7.04$^{***}$ & [+4.55, +9.54] & $<$0.001 \\
 & Citation Count & -12.82$^{***}$ & [-15.83, -9.81] & $<$0.001 \\
 & Citation Age & -0.59\phantom{$^{***}$} & [-1.84, +0.67] & 0.359 \\
 & Response Length & -0.04\phantom{$^{***}$} & [-1.50, +1.43] & 0.960 \\
 & Response Content & +9.17$^{***}$ & [+6.33, +12.01] & $<$0.001 \\
\bottomrule
\end{tabular}
\caption{Estimated fixed effects on human votes and LLM judgments for \textbf{SciArena}, reported as changes in win rate (percentage points) for a one standard deviation increase in each standardized predictor. This table reports the exact estimates underlying Figure~\ref{fig:fixed-effects-sciarena}. Significance: $^{***}p<0.001$; $^{**}p<0.01$; $^{*}p<0.05$.}
\label{tab:sciarena-estimates}
\end{table*}
\begin{table*}[t]
\centering
% \small
% \setlength{\tabcolsep}{5pt}
\begin{tabular}{lllrr}
\toprule
Judge & Predictor & $\Delta$Win Rate & 95\% CI & $p$ \\
\midrule
Human & Citation Diversity & +11.65$^{***}$ & [+5.31, +17.99] & $<$0.001 \\
 & Citation Count & -6.47$^{*}$ & [-12.88, -0.05] & 0.048 \\
 & Citation Mismatch & -10.08$^{**}$ & [-17.01, -3.15] & 0.004 \\
 & Response Length & +21.41$^{***}$ & [+13.86, +28.96] & $<$0.001 \\
 & Response Content & -0.22\phantom{$^{***}$} & [-8.66, +8.22] & 0.959 \\
\midrule
Llama & Citation Diversity & +7.76$^{***}$ & [+3.99, +11.54] & $<$0.001 \\
 & Citation Count & +0.67\phantom{$^{***}$} & [-3.05, +4.39] & 0.722 \\
 & Citation Mismatch & +0.07\phantom{$^{***}$} & [-3.82, +3.95] & 0.973 \\
 & Response Length & +9.10$^{***}$ & [+4.36, +13.84] & $<$0.001 \\
 & Response Content & +1.07\phantom{$^{***}$} & [-3.68, +5.83] & 0.659 \\
\midrule
Gemma & Citation Diversity & +8.82$^{***}$ & [+4.62, +13.03] & $<$0.001 \\
 & Citation Count & +1.53\phantom{$^{***}$} & [-2.66, +5.73] & 0.474 \\
 & Citation Mismatch & -2.33\phantom{$^{***}$} & [-5.99, +1.32] & 0.211 \\
 & Response Length & +11.16$^{***}$ & [+6.05, +16.27] & $<$0.001 \\
 & Response Content & +2.76\phantom{$^{***}$} & [-2.53, +8.04] & 0.307 \\
\midrule
Qwen & Citation Diversity & +4.35\phantom{$^{***}$} & [-0.30, +8.99] & 0.067 \\
 & Citation Count & -0.66\phantom{$^{***}$} & [-5.35, +4.03] & 0.781 \\
 & Citation Mismatch & +0.44\phantom{$^{***}$} & [-3.63, +4.52] & 0.832 \\
 & Response Length & +5.87$^{*}$ & [+0.24, +11.49] & 0.041 \\
 & Response Content & +7.23$^{*}$ & [+1.53, +12.92] & 0.013 \\
\midrule
Skywork & Citation Diversity & +0.34\phantom{$^{***}$} & [-2.85, +3.53] & 0.833 \\
 & Citation Count & -2.25\phantom{$^{***}$} & [-5.66, +1.16] & 0.195 \\
 & Citation Mismatch & +1.18\phantom{$^{***}$} & [-1.22, +3.58] & 0.336 \\
 & Response Length & +1.95\phantom{$^{***}$} & [-1.97, +5.88] & 0.330 \\
 & Response Content & +0.66\phantom{$^{***}$} & [-3.20, +4.53] & 0.737 \\
\bottomrule
\end{tabular}
\caption{Estimated fixed effects on human votes and LLM judgments for \textbf{ResearchQA}, reported as changes in win rate (percentage points) for a one standard deviation increase in each standardized predictor. This table reports the exact estimates underlying Figure~\ref{fig:fixed-effects-researchqa}. Significance: $^{***}p<0.001$; $^{**}p<0.01$; $^{*}p<0.05$.}
\label{tab:researchqa-estimates}
\end{table*}

\section{Usage of Large Language Models}
We use LLMs to support and refine the writing of our work. We do not rely on them to generate content or sentences from scratch and take full responsibility for the final outcome.
In addition, we use AI to help brainstorm figure visualizations and assist with coding when necessary.

\end{document}